%% file: main.tex
\documentclass{ceurart}
\usepackage{booktabs}
\usepackage{tikz}
\usepackage{tabularx}
\usepackage{amssymb}
\usepackage{graphicx}
\usepackage{xurl}
\usepackage{algorithm}
\usepackage{algpseudocode}

\makeatletter\renewcommand{\ALG@beginalgorithmic}{\footnotesize\linespread{0.92}\selectfont}
\makeatother
\input{generated/numbers.tex}

\makeatletter
\newcommand{\URLprefix}[2]{\in@{dx.doi.org}{#2}%
  \ifin@\expandafter\ceur@skipdot\else URL: \url{#2}\fi}
\def\ceur@skipdot{\@ifnextchar.\ceur@gobbledot\relax}
\def\ceur@gobbledot.{\ignorespaces}
\makeatother

\begin{document}

\makeatletter\def\@copyrightLine{}\makeatother
\conference{Preprint, November 2026}

\title{One Scale, Four Texts: Gaps, Overlaps and Silent Repairs in Published Grading Schemes}

\author[1]{Elisa Rovai}[]
\fnmark[1]
\address[1]{University of Padua, Padua, Italy}
\author[2]{Fabio Rovai}[email=fabio@thetesseractacademy.com]
\cormark[1]
\fnmark[1]
\address[2]{The Tesseract Academy, London, United Kingdom}
\cortext[1]{Corresponding author.}
\fntext[1]{These authors contributed equally.}

\begin{abstract}
Category systems often reach their users as text, and readers apply them with tacit repairs. We ask what such a text determines once the repairs are removed, using diabetic retinopathy grading, where one international scale circulates in several documents. We transcribe four texts (the English screening definitions of 2025, the international scale as tabulated in 2017 and as published in 2003, and an Italian guideline of 2015) into one rule language over \NVars{} finding variables, and make every point a text leaves open a named precisification. Exhaustive enumeration of \SpaceSize{} finding profiles, re-derived by an SMT solver and spot-checked by an OWL reasoner, shows that every text that assigns actions leaves some eyes without a category under every reading; that the 2017 tabulation has a proliferative class that is empty when read literally and a macular gap that the 2003 text lacks, so that no function maps the 2003 grades onto the 2017 grades; and that on image findings alone the English and Italian texts order some pairs of eyes in opposite ways. On \TrNAll{} annotated photographs from three public datasets, with retinal thickening assumed absent, the English text and the 2017 tabulation disagree on referral for \FqDdrENICdisPct\% of the eyes of one dataset (DDR) on which both decide, and in a pre-registered experiment classifiers trained on each text's labels side with their own text. Thickening cannot be seen on a colour photograph; left free, every disagreement involving England becomes undecidable, so it rests on what the camera cannot show. In a second pre-registered study, language models that answered without deliberating repaired most gap and overlap items even when told to read literally, while \SonnetName{} and \OpusName{} reported the defects far more often; within two models, deliberation reduced silent repair without removing it.
\end{abstract}


\maketitle

\section{Introduction}

Much institutional categorisation is written down before it is applied. A grading scheme or a triage table is a text, and whoever applies it reads it with conventions the text does not state: take the worst category when two apply, choose the nearest sensible one when none does, ignore a literal reading that would be absurd. These repairs are silent, and they stop working when a text is executed rather than read: coded into software, used to define labels for machine learning, matched against another text in a crosswalk that pools data, or handed to a language model. This paper asks what a published category system determines on its own. We take the texts at their word, turn every open point into an explicit choice among admissible readings, and ask which defects survive every choice.

Diabetic retinopathy grading suits the question: its categories are explicit rules over a shared vocabulary of lesions, several texts claim to state the same scale, and annotated images exist. We use the English NHS Diabetic Eye Screening definitions of December 2025~\cite{nhs_fbg2025,nhs_referable2025}, the International Clinical Diabetic Retinopathy (ICDR) scale as tabulated by the International Council of Ophthalmology (ICO) in 2017~\cite{ico2017} and as published in 2003~\cite{wilkinson2003}, and Table~3 of the Italian guideline of 2015~\cite{italy2015}. Our claims are about texts, never about medicine or patients. A category is correct in our sense if the text licenses it, and every category and reading in the transcription is linked to a verbatim quote from its source, so each claim can be checked by comparing a rule with the words it came from, without clinical judgement.

The question is not whether guidelines leave room for clinical judgement. A text may leave a decision to the grader on purpose, and we keep every such opening as a named reading rather than resolve it. A gap is something else: an eye that no category covers under any admissible reading, so that the text applied as written returns nothing. An overlap returns two categories with different actions, and an order inversion means that two texts rank the same pair of eyes in opposite ways. A grader absorbs the first two without noticing, but software, a label generator or a language model has to return something definite, and what it returns is then decided by the implementation, not by the text. Knowing where a text is silent or contradictory, and whether two texts that share a name define the same categories, is a precondition for pooling labels across them.

We make four contributions. Section~\ref{sec:formal} gives a supervaluational treatment of competing readings and texts, with an explicit criterion for which readings are admissible. Section~\ref{sec:defects} reports the gaps, overlaps, empty classes and order inversions that survive every reading, including a change between the 2003 and 2017 versions of the international scale that no relabelling can undo. Section~\ref{sec:real} measures how often annotated photographs fall into these defects, shows how much of that depends on an assumption about retinal thickening, and tests in a pre-registered experiment whether classifiers trained on labels from different texts differ. Section~\ref{sec:llm} reports a pre-registered test of whether language models repair defective categories silently. Section~\ref{sec:align} aligns the transcription with Basic Formal Ontology and standard terminologies, and Section~\ref{sec:artefact} describes how every claim is checked: \NMakeGates{} build gates, \NClaims{} assertions over the result files, three reasoning engines, and a sabotage suite of \NSabotage{} cases that tests whether each gate can fail. Appendix~\ref{app:algs} summarises each procedure as pseudocode (\NAlgorithms{} algorithms).

\section{Related work}

Psychology distinguishes categories with graded structure organised around prototypes~\cite{rosch1975} from rule-based categories learned by explicit reasoning, supported by distinct systems~\cite{ashby2005}; grading texts take the rule-based form. For the vague terms in the texts (``isolated'', ``prominent'', ``approaching the center'') we use supervaluation~\cite{fine1975}: a statement is determinately true if it holds under every admissible precisification. Standpoint semantics refines this idea for vague terms in a formal ontology~\cite{bennett2008}, and standpoint logic makes perspectives on a shared vocabulary first-class~\cite{gomezalvarez2021}, also for OWL-style languages~\cite{gomezalvarez2025}. In Bennett's setting standpoints are precisifications of one vague term; in ours each text is a standpoint whose readings are its precisifications. Fuzzy ontologies~\cite{bobillo2011} do not fit, since the texts assign crisp grades and a gap is not a matter of degree.

Checking guidelines for completeness and consistency is an old idea. Shiffman and Greenes enumerated every combination of the variables in an immunisation guideline, removed the impossible ones, and reported the combinations no rule covered and the identical conditions that received different actions~\cite{shiffman1994}; augmented decision tables made such checks part of guideline representation~\cite{shiffman1997}. Later work formalised medical protocols in a plan language and verified quality properties with a theorem prover, finding ambiguity and incompleteness~\cite{tenteije2006}, or verified guideline task networks against medical background knowledge~\cite{hommersom2007}; Peleg reviews computer-interpretable guidelines and their verification~\cite{peleg2013}. Our gap and overlap detection is the decision-table check of~\cite{shiffman1994} applied to grading texts. What we add is supervaluation over explicit readings, so that a reported defect does not depend on one interpretation of vague wording; the comparison of texts that claim to state the same scale, including whether one version can be mapped onto another; and three engines whose agreement is itself checked. For diabetic retinopathy, OWL disease models have been built from SNOMED CT and clinical sources~\cite{arguello2017}, and the variety of grading systems behind public image datasets has been described as a ground-truth problem for machine learning~\cite{nakayama2022}; neither analyses the grading texts as category systems.

Guidelines have long been modelled as conditionals, imperatives and definitions~\cite{shiffman2000}; appraisal treats decidability as a critical property~\cite{shiffman2005}, and readers interpret guideline wording with wide variability~\cite{lomotan2010}. Our gaps and overlaps are decidability failures in definitions, found mechanically, and differ from grader variability under one scheme~\cite{krause2018}: they remain when every text is applied perfectly. In ontology alignment, the correctness of a correspondence depends on the semantics of both sides and on the other correspondences~\cite{meilicke2009}, and mappings between medical terminologies contain errors visible only as unintended logical consequences~\cite{jimenezruiz2011}; a crosswalk between grading texts that is not a function is our analogue. For the upper-level placement we follow Basic Formal Ontology (BFO)~\cite{arp2015} and the Information Artifact Ontology's treatment of data about an entity~\cite{ceusters2012}. Language models have been evaluated on learning ontologies from text~\cite{babaeigiglou2023}, framed as performing ontology-like functions~\cite{neuhaus2023}, and found to confuse similar rules~\cite{zhou2025}; we ask what they do with a defective category system.

\section{Texts, readings and determinacy}
\label{sec:formal}

\begin{table}[t]
\caption{Clinical terms used in the paper. The definitions are those a reader needs to follow the rules; none of our claims depends on clinical knowledge beyond them.}
\label{tab:terms}
\footnotesize
\begin{tabularx}{\linewidth}{@{}lX@{}}
\toprule
term & meaning \\
\midrule
DR; NPDR, PDR & diabetic retinopathy, damage to the retinal vessels caused by diabetes; its non-proliferative and proliferative stages. Proliferative disease is marked by new vessels or bleeding from them. \\
DME (DMO) & diabetic macular oedema, thickening of the retina at the macula; the English text writes DMO. \\
microaneurysm (MA) & a small red dot, the bulge of a capillary wall, usually the first visible sign of DR. \\
haemorrhage & a bleed inside the retina (dot, blot or flame); a preretinal or vitreous haemorrhage lies in front of the retina or in the vitreous gel. \\
hard exudate & a yellow deposit of lipid leaked from damaged vessels. \\
cotton wool spot & a pale patch caused by local loss of blood supply to the nerve-fibre layer. \\
venous beading, venous loop & irregular calibre of a retinal vein, and a vein that doubles back on itself. \\
IRMA & intraretinal microvascular abnormalities: dilated or newly formed capillaries within the retina. \\
new vessels (NV) & abnormal vessels growing on the optic disc or elsewhere on the retina. \\
fovea, macula & the centre of sharp vision and the region around it. \\
optic disc; DD & where the optic nerve leaves the eye; its diameter, the disc diameter (DD), is the unit of distance on the retina. \\
retinal thickening & swelling of the retina, judged by stereo examination or optical coherence tomography; a single colour photograph does not show it. \\
4-2-1 rule & the criteria for severe NPDR: many haemorrhages in all four quadrants, venous beading in two or more, or IRMA in one or more. \\
photograph 8A & a standard photograph of the Early Treatment Diabetic Retinopathy Study that sets the threshold for marked IRMA. \\
R0--R3A; M0, M1 & England's retinopathy grades (none, background, pre-proliferative at lower or higher risk, active proliferative) and maculopathy grades (absent, present). \\
action & what a grade or table row prescribes, on the scale rescreen, recall, refer, urgent referral. \\
\bottomrule
\end{tabularx}
\end{table}

\paragraph{Sources and findings.} Table~\ref{tab:terms} explains the clinical terms. England grades retinopathy (R0 to R3) and maculopathy (M0, M1) separately~\cite{nhs_fbg2025,nhs_referable2025}. For ICO 2017 we use the source's own Table~1 and its high-resource referral Table~2a~\cite{ico2017}. The 2003 text~\cite{wilkinson2003} was read in its 2016 reproduction by the American Academy of Ophthalmology, checked against the October 2002 consensus tables; it has grades but no referral table. The Italian Table~3 maps findings directly to a presumptive diagnosis and an action~\cite{italy2015}; we transcribe its rows up to proliferative disease, leaving out high-risk proliferative and advanced disease. Treatment history is out of scope. An eye is described by \NVars{} ordinal variables binned at every threshold any text uses (lesion counts, quadrants above the haemorrhage thresholds, venous beading, IRMA, new vessels, preretinal and vitreous haemorrhage, distances from the fovea). Four variables are not visible on a colour photograph: reduced acuity, a known non-oedema cause of it, retinal thickening, and a grader's report of a finding they cannot interpret with confidence. \NConstraints{} constraints exclude impossible profiles. Since every rule compares a variable with one of these thresholds, two eyes in the same bins get the same category from every text (Algorithm~\ref{alg:enum}). The space has \SpaceSize{} profiles.

\paragraph{Readings.} Each text becomes a set of Boolean rules over threshold tests, and every open point a parameter. England has \NPEN{} readings (whether the maculopathy exemption overrides the general route to M1, how the stereo route applies, and what ``isolated cotton wool spots'' requires). ICO 2017 has \NPIC{}: ``$\geq 20$'' haemorrhages per quadrant as printed or ``more than 20'', beading ``in 2 quadrants'' as exactly or at least two, IRMA as any or prominent, and the proliferative row read literally, by the severe criteria only, or as any proliferative sign. ICDR 2003 has \NPIThree{} (where ``no signs of proliferative retinopathy'' attaches, whether ``More than just microaneurysms'' requires microaneurysms, the ``posterior pole'' and ``approaching'' the centre), and Italy \NPIT{}, chiefly the numeric meaning of ``isolati'' (isolated) and ``multipli'' (multiple). Outcomes map onto a common action scale: rescreen (12 months or later), recall (6 to 12 months or surveillance), refer, and urgent referral.

\paragraph{Which readings are admissible.} A reading is admissible if every choice in it is a rule the quoted wording supports, and each value of each reading parameter is linked to the quote that licenses it (Section~\ref{sec:artefact}). We add a parameter wherever the wording admits more than one rule: a vague term (``isolated'', ``prominent''), a printed threshold that differs from other statements of the same scale (``$\geq 20$'' against ``more than 20''), a clause whose scope is unclear, or a precedence between rules that the text does not state. We add no reading that needs clinical knowledge the text does not state, so the set describes the text, not good practice. The set cannot be shown complete, and the two kinds of claim respond differently to a reading we missed: a defect under some reading survives any added reading, while a defect under every reading would be removed by an admissible reading that avoids it. The parameters are listed above, so a reader who disputes a defect under every reading can name the reading that removes it.

\paragraph{Determinacy.} Let $P_S$ be the readings of text $S$ and $a^p_S(x)$ the action for profile $x$ under $p$: the worst action among the grades or rows that apply, undefined if none applies. The action is \emph{determinate} if it is defined and the same for every $p\in P_S$, which is supertruth in Fine's sense. Two texts \emph{structurally disagree} at $x$ if both actions are determinate and differ, so that no admissible reading of either removes the disagreement. Two texts \emph{invert} on $x,y$ if all four actions are determinate, $a_S(x)>a_S(y)$ and $a_T(x)<a_T(y)$ (Algorithm~\ref{alg:det}). Disagreement without inversion is compatible with a monotone crosswalk that only moves thresholds; an inversion rules every monotone crosswalk out. In standpoint terms, a structural disagreement holds between standpoints whatever sharpening each adopts.

\paragraph{A worked example.} Take an eye whose only finding is a hard exudate outside the macula. In the English text, R0 requires that ``no features of diabetic retinopathy are present'', which fails because an exudate is such a feature, and its exception for isolated cotton wool spots needs a cotton wool spot. R1 admits an exudate only ``in the presence of other features of DR'', and R2L, R2H and R3A each need a lesion the eye lacks. The \NPEN{} English readings vary only the cotton wool spot clause and the maculopathy rules, so under all of them the retinopathy axis has no grade, while the maculopathy axis gives M0 because the exudate lies outside the macula. The action, the worst over the two axes, is therefore undefined under every reading: a gap that survives supervaluation. This eye is the English witness in Figure~\ref{fig:gaps}.

\section{What the texts determine}
\label{sec:defects}

\begin{figure}[t]
\centering
\begin{minipage}[c]{0.47\linewidth}\centering
\includegraphics[width=\linewidth]{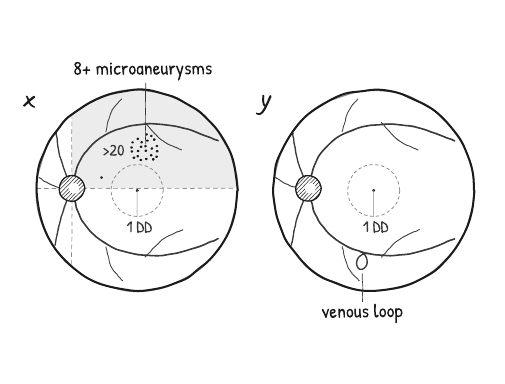}
\end{minipage}\hfill
\begin{minipage}[c]{0.5\linewidth}\centering
\resizebox{\linewidth}{!}{\input{generated/fig_inversion.tex}}
\end{minipage}
\caption{An order inversion between England and Italy on image findings alone. Each line joins the actions one eye receives from the three texts, each determinate under every reading; the lines cross between England and Italy. \protect\input{generated/fig_inversion_caption.tex}}
\label{fig:inversion}
\end{figure}

\subsection{Gaps}

\begin{figure}[t]
\centering
\includegraphics[width=0.74\linewidth]{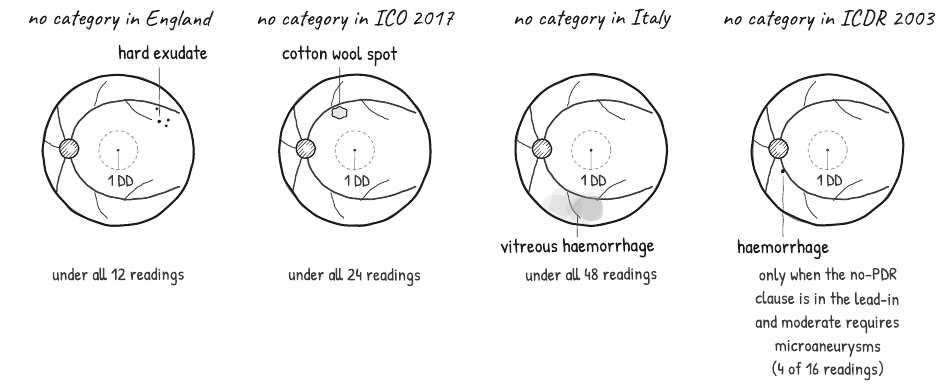}
\caption{A minimal gap in each text: an eye that no category of the text covers. \protect\input{generated/fig_gaps_caption.tex}}
\label{fig:gaps}
\end{figure}

Every text that assigns actions has a gap under all of its readings (Figure~\ref{fig:gaps}). England gives no retinopathy grade to an eye whose only finding is a hard exudate, at any distance from the fovea: R0 requires that ``no features of diabetic retinopathy are present'' and R1 admits ``any exudate in the presence of other features of DR''~\cite{nhs_fbg2025}. In the ICO table every non-proliferative grade above ``No apparent DR'' requires microaneurysms (``Microaneurysms only'', ``Microaneurysms and other signs''), so haemorrhages, cotton wool spots, IRMA, venous beading or a venous loop without microaneurysms receive no grade under all \NPIC{} readings. New vessels alone receive none under exactly the readings that tie proliferative disease to severe NPDR. Read literally, the proliferative class is empty, as Z3 proves: it requires severe NPDR, listed ``with any of the following: \ldots{} and no signs of proliferative retinopathy'', while proliferative disease requires a proliferative sign. The Italian table has a gap under all \NPIT{} readings: vitreous haemorrhage alone fits no row, since the urgent row lists ``neovasi della papilla ottica e/o della retina; emorragie preretiniche'' (new vessels of the optic disc and/or the retina; preretinal haemorrhages) and no row mentions vitreous haemorrhage. Other Italian gaps depend on the reading: two to seven haemorrhages without microaneurysms fit no row exactly when ``isolated'' means one and ``multiple'' eight or more. The 2003 text, by contrast, has gaps only under some of its readings. Each witness has the fewest findings away from a healthy eye (Algorithm~\ref{alg:witness}).

\subsection{One scale, two texts}

\begin{figure}[t]
\centering
\includegraphics[width=0.44\linewidth]{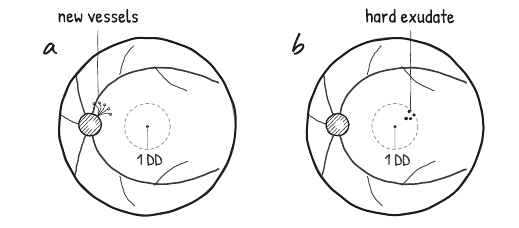}\par\medskip
\small
\input{generated/fig_versions.tex}
\caption{The same eyes under the two versions of the international scale. Each cell gives the grade and, below it, the readings under which the text assigns it. \protect\input{generated/fig_versions_caption.tex}}
\label{fig:versions}
\end{figure}

The two versions of the international scale differ as category systems (Figure~\ref{fig:versions}). The empty proliferative class belongs to 2017 alone: it is empty under exactly the literal ICO readings, while no 2003 grade is empty under any of the \NPIThree{} readings, the 2003 proliferative row asking only for ``One or both of the following'' signs. The macular gap is also new. ICO 2017 defines ``No DME'' as ``No retinal thickening or hard exudates in the macula'' but both oedema grades by thickening alone, so exudates in the macula without thickening fall between them under every ICO reading; the 2003 text, where each level reads ``retinal thickening or hard exudates'', has no macular gap under any reading. The retinopathy gap caused by the microaneurysm requirement holds under every ICO reading, but in 2003 only when ``More than just microaneurysms'' requires microaneurysms and the no-proliferative clause governs the whole list. Overlaps run the other way: 2003 retinopathy grades overlap under every reading, 2017 grades only when the proliferative row is not read literally. Relabelling 2003 grades as 2017 grades is not a function: for both reference readings of ICO 2017 and every 2003 reading, some 2003 grade meets more than one 2017 grade.

\subsection{Overlaps}

\begin{figure}[t]
\centering
\begin{minipage}[c]{0.24\linewidth}\centering
\includegraphics[width=\linewidth]{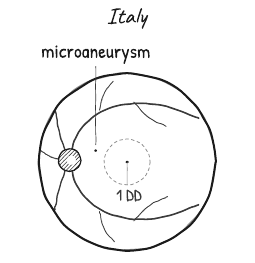}
\end{minipage}\hfill
\begin{minipage}[c]{0.73\linewidth}\centering\small
\resizebox{\linewidth}{!}{\input{generated/fig_overlap.tex}}
\end{minipage}
\caption{An overlap in the Italian table: two rows with different actions apply to the same eye under every reading. \protect\input{generated/fig_overlap_caption.tex}}
\label{fig:overlap}
\end{figure}

England's cotton wool spot with an exudate is both R0 and R1 only under the reading in which the absence clause defines ``isolated'', and both grades mean rescreening. The maculopathy overlap matters more: without stereo examination, microaneurysms within one disc diameter of the fovea with reduced vision of known non-oedema cause are M1 by the general rule and M0 by the exemption, and the eye is referred or rescreened according to a precedence the text does not state. In the ICO referral table a single microaneurysm matches both ``No apparent DR, mild nonproliferative DR and no DME'' (one to two years) and ``Mild nonproliferative DR'' (six to twelve months) under every reading. In the \ICOvlDRReadings{} readings that do not take the proliferative row literally, microaneurysms with new vessels and IRMA at least as severe as photograph 8a are both moderate and proliferative, because moderate means ``less than severe'' and severe excludes proliferative signs. In the Italian table a single microaneurysm away from the fovea fits both the mild row (``microaneurismi isolati'', isolated microaneurysms, twelve months) and the moderate row (``emorragie isolate e/o microaneurismi'', isolated haemorrhages and/or microaneurysms, six to twelve months) under every reading (Figure~\ref{fig:overlap}). One Italian row, ``qualsiasi altro reperto che l'osservatore non si senta in grado di interpretare con sicurezza'' (any other finding that the observer does not feel able to interpret with confidence), assigns a disease stage to the grader's own uncertainty: a category about the reader, not the retina.

\subsection{Order inversions}

On the \ImageSpaceSize{} profiles without non-image findings, every pair of action texts structurally disagrees about referral somewhere. England inverts against Italy (Figure~\ref{fig:inversion}): take eye $x$ with \FigXDesc{}, and eye $y$ with \FigYDesc{} alone. England assigns $x$ \FigXEN{} and $y$ \FigYEN{}, Italy assigns $x$ \FigXIT{} and $y$ \FigYIT{}, and ICO 2017 assigns \FigXIC{} and \FigYIC{}, each under all of its readings. England also inverts against ICO 2017: microaneurysms alone exceeding 20 in every quadrant are referred in England and recalled by ICO, while microaneurysms with a cotton wool spot are rescreened in England and referred by ICO. ICO 2017 and Italy never invert on image findings, so theirs is a disagreement of thresholds. Once acuity, thickening or grader uncertainty are allowed, all three pairs invert, and under every reading of every action text two eyes with identical image findings receive different actions, so no photograph-only model can reproduce any of the texts exactly.

\section{Real eyes}
\label{sec:real}

The defects above belong to the texts; we now ask how often annotated photographs fall into them and whether labels from different texts produce different classifiers. We use three public sets with lesion annotations. IDRiD~\cite{porwal2018}, released under the Creative Commons Attribution 4.0 licence, has a segmentation set of \FqIdN{} photographs with pixel masks for each lesion type, and 516 photographs (413 for training, 103 for testing) that carry disease grades and, in the localisation part, marked disc centre and fovea; we count lesions on the segmentation set and train the detector below on the localisation marks. The lesion-segmentation subset of DDR~\cite{li2019} has \FqDdrN{} images under the Creative Commons Attribution-NonCommercial-ShareAlike 4.0 licence; our copy was checked file by file against the official release. DIARETDB1~\cite{kauppi2007} has \FqDdbN{} images under a restricted academic licence, and we take its findings as the strict-majority consensus of its experts. Lesion counts come from the annotations, and findings they do not record are taken as absent, with normal acuity, no retinal thickening and a confident grader (the \emph{closed completion}). The images come from hospitals and were chosen for annotation, so they are enriched for disease (in DDR, \FqPrevMA\% have microaneurysms, \FqPrevHE\% haemorrhages and \FqPrevEX\% hard exudates), and the shares below describe these sets, not a screening population.

\paragraph{Positions from a validated detector.} Distances from the fovea, the macular radius and the quadrant counts need the fovea, the disc centre and the disc diameter (DD), which only IDRiD marks, and only in part, so we trained a detector. A U-Net~\cite{ronneberger2015} with a \DetArch{} encoder~\cite{he2016} pre-trained on ImageNet predicts two heatmaps, one for the disc centre and one for the fovea. We chose an encoder-decoder with a pre-trained encoder because it predicts dense maps at the input resolution and can be trained from a few hundred images, and we fixed it before training. The photograph is scaled so that the long side of the fundus is 512 pixels; each target is a Gaussian of 5 pixels' standard deviation at the marked point, the loss is the cross-entropy between the target and a softmax over the map, and a position is the argmax of the predicted map refined by a weighted centroid around it. The model was trained on 287 localisation training images with flips, scaling, rotation, translation, and colour and gamma jitter (AdamW, learning rate $3\times 10^{-4}$, one-cycle schedule, 16 epochs), the epoch was chosen on 125 others, and one training image that duplicates a test image was dropped. A second U-Net with the same encoder segments the disc on a 256-pixel crop around the predicted disc centre, trained with binary cross-entropy and Dice loss in five-fold cross-validation over the \DetNMasks{} disc masks of the segmentation set; the disc diameter is that of the circle with the area of the largest predicted component. The acceptance gate was written before training and its SHA-256 recorded (prefix \DetGateSha); evaluated once on the \DetNTest{} localisation test images, the detector passed all four conditions: median fovea error \DetFovMed{} DD (required at most \DetFovMedThr), 95th percentile \DetFovP{} DD (at most \DetFovPThr), median disc-centre error \DetDiscMed{} DD (at most \DetDiscThr), and median diameter error \DetDDMed\% in cross-validation over \DetNMasks{} disc masks (at most \DetDDThr\%). Each position variable takes the set of bins compatible with error margins set at the 95th percentiles of a calibration set (fovea \DetEFov{} DD, disc centre \DetEDisc{} DD, diameter \DetEDD\%), and a set-valued variable is quantified like an unrecorded one: the lower bound of a defect counts the eyes that have it for every value in the set, the upper bound those that have it for some. On the \DetCovN{} images with a true fovea mark, the true bin lay in the set in all \DetCovChecks{} checks, while without margins \DetCovNoMarginMiss{} checks miss, so the coverage test can fail. Against the disc annotations of DDR and DIARETDB1 the diameter error is larger (medians \DetDdrDDMed\% and \DetDdbDDMed\%, 95th percentiles \DetDdrDDP\% and \DetDdbDDP\%), so for those sets the disc margins are widened to each set's own 95th percentile; the fovea margin keeps its IDRiD value, since neither set marks the fovea. For IDRiD we widened the fovea margin after testing, to the test 95th percentile; a wider margin can only add bins, so this loosens the bounds and never tightens them.

\begin{table}[t]
\caption{Share (\%) of eyes in each defect class, with detected positions and the closed completion. A range gives the lower and upper bound over the set-valued positions (and, in the last row, over retinal thickening); ``every'' and ``some'' quantify over the readings of the text. The last row repeats the ICO gap with thickening left free instead of assumed absent.}
\label{tab:real}
\footnotesize\centering
\begin{tabular}{@{}lllccc@{}}
\toprule
text & defect & readings & IDRiD (\FqIdN) & DDR (\FqDdrN) & DIARETDB1 (\FqDdbN) \\
\midrule
\input{generated/tab_realeyes.tex}
\end{tabular}
\end{table}

\paragraph{How often real eyes fall into each defect.} Of the gaps and overlaps that hold under every reading, the ICO gap is the commonest in every sample (Table~\ref{tab:real}). On IDRiD it is the macular gap of Section~\ref{sec:defects}: exudates in the macula without thickening, which the closed completion supplies because a colour photograph cannot show thickening. On the \FqIdSixty{} IDRiD images with a fovea mark it holds for every image under the closed completion and for none once thickening is left free. In DDR the macular gap covers \FqICmacDdrLo{} to \FqICmacDdrHi{} eyes, and the microaneurysm requirement adds \FqICnoMA{}, exactly the eyes whose lesions include no marked microaneurysm, so this part depends on how completely the DDR annotators marked microaneurysms. The English gap is rare (\FqENgapDdr{} DDR eyes, \FqENgapDdrPct\%, 95\% Wilson interval \FqENgapDdrLo--\FqENgapDdrHi\%, and \FqENgapDdb{} in DIARETDB1), and each such eye has exudates and no other annotated lesion, like the witness in Figure~\ref{fig:gaps}. Nearly all English overlaps on IDRiD and DDR pair R1 (rescreen) with R2L (recall), \FqENovDdrLo{} to \FqENovDdrHi{} DDR eyes, and worst-grade precedence settles them; the overlap of R0 and R1 discussed above occurs on one DDR and one DIARETDB1 eye, under some readings only. Every ICO overlap under every reading matches the first two rows of the referral table, as the single microaneurysm does (\FqICovDdr{} DDR and \FqICovDdb{} DIARETDB1 eyes). Italian overlaps under every reading mostly join the moderate and severe rows (\FqITovDdrLo{} to \FqITovDdrHi{} DDR eyes), and the 2003 grade changes with the reading on almost every eye. No sample has a gap under every reading for the 2003 text or the Italian table; the Italian one needs a vitreous haemorrhage, which no dataset annotates. Under the closed completion, where England and ICO both decide referral under every reading and every completion, on \FqDdrENICbothN{} DDR eyes, they disagree on \FqDdrENICdisN{} (\FqDdrENICdisPct\%, \FqDdrENICdisLo--\FqDdrENICdisHi\%); England and Italy disagree on \FqDdrENITdisN{} of \FqDdrENITbothN, ICO and Italy on \FqDdrICITdisN{} of \FqDdrICITbothN. On the \IdMatched{} IDRiD images with a fovea mark, England and ICO disagree on \ENICdis{}, and neither survives in 95\% of \NDraws{} draws that resample counts with inter-expert ratios from DIARETDB1. The IDRiD annotations have defects of their own: on \DmeContra{} images the oedema-risk label contradicts the exudate masks, and one photograph appears twice with different grades.

\paragraph{What the absence of thickening contributes.} The closed completion assumes that the retina is not thickened, which a colour photograph cannot confirm, and both the ICO oedema grades and the English stereo readings turn on thickening. We therefore repeated the analysis with thickening left free and every other variable as before. The gaps and overlaps of the English, 2003 and Italian texts do not change. The ICO gap under every reading falls to \TfICgapId\% on IDRiD, \TfICgapDdr\% on DDR (the \TfICgapDdrN{} eyes without a marked microaneurysm) and \TfICgapDdb\% on DIARETDB1 (last row of Table~\ref{tab:real}). The referral disagreements involving England disappear: on the \TfDdrENICbothN{} DDR eyes on which England and ICO both decide with thickening free they always agree, and so do England and Italy on \TfDdrENITbothN{}, while ICO and Italy keep their \TfDdrICITdisN{} disagreements out of \TfDdrICITbothN. On these photographs, then, England departs from the other two texts only on eyes whose macula is assumed not thickened. This is the dependence on non-image findings of Section~\ref{sec:defects} seen in data: the photograph alone does not decide the English referral on these eyes, and the disagreement lies in what the camera cannot show.

\begin{table}[t]
\caption{Accuracy (\%, 95\% Wilson interval) of the classifier trained on each text's referral labels (rows) against each text's labels (columns), pooled over the held-out folds of the \TrN{} images that all three texts label.}
\label{tab:train}
\footnotesize\centering
\begin{tabular}{@{}lccc@{}}
\toprule
trained on $\downarrow$ / evaluated against $\rightarrow$ & England & ICO 2017 & Italy \\
\midrule
\input{generated/tab_train.tex}
\end{tabular}
\end{table}

\paragraph{Labels from different texts produce different models.} This experiment was pre-registered at \TrPreregUTC{} UTC (SHA-256 prefix \TrPreregSha), before any label was computed. An image has a referral label under a text when the decision is the same under every reading and every profile its annotations allow; \TrN{} of the \TrNAll{} images (\TrNId{} IDRiD, \TrNDdr{} DDR, \TrNDdb{} DIARETDB1) have one under all three action texts. On this set the labels are nested: \TrTripNone{} images are referable under no text, \TrTripIC{} under ICO only, \TrTripICIT{} under ICO and Italy and \TrTripAll{} under all three, so the texts cut one ordering at different places, England referring fewest (\TrPrevEN\%, against \TrPrevIC\% for ICO and \TrPrevIT\% for Italy). We trained one \TrArch{} classifier per text on the photographs, with \TrFolds-fold cross-validation over photo groups and the same folds, recipe and seeds for every text, and evaluated each model against the labels of every text (Table~\ref{tab:train}). The confirmatory hypothesis was that a model trained on text A agrees less with the labels of text B than with those of A; an exact McNemar test compares the two for each ordered pair, and its discordant images are exactly those on which the labels differ. All \TrTestable{} testable pairs, the four involving England, support it (Holm-adjusted $p$ from $\TrHolmMin$ to \TrHolmMax), so the registered verdict is ``\TrEightVerdict''. ICO and Italy disagree on only \TrUnderN{} images, fewer than the registered minimum of \TrUnderThr, so their two pairs are untestable, and the ICO model is as accurate against Italian labels as against its own. Where labels differ, a model sides with its own text on \TrSideLo\% to \TrSideHi\% of the images, and the exploratory comparison of AUROC was \TrTenVerdict{} for the same four pairs. A registered bound, that accuracy against B cannot exceed the agreement of the labels plus the error against A, holds by the triangle inequality for Hamming distance and serves as a consistency check. One deviation was logged before any label existed: the rule that pooled the DIARETDB1 experts' positions left \TrDevImgs{} images without a profile and was restricted to lesion types present in the consensus. The common set holds the images that every text decides, under half of all images, so the divergence involving England appears even on clear-cut eyes. The labels use the closed completion, and the English divergence depends on it: with thickening left free, every one of the \TrTfENICdis{} images on which the English and ICO labels differ, and every one of the \TrTfENITdis{} on which the English and Italian labels differ, loses its English label, because the English decision on it turns on whether the macula is thickened; the \TrUnderN{} ICO--Italian disagreements remain. The classifiers therefore learn each text's decision on eyes assumed to have no thickening, and the English divergence they reproduce concerns what the photograph does not show. This analysis was made after the experiment and changes none of its registered results.

\section{Ontological placement and vocabulary alignment}
\label{sec:align}

An alignment module imports the OWL~2 transcription unchanged with the 2020 release of BFO and the Information Artifact Ontology (IAO), adding only subclass, restriction and annotation axioms. The eye and its lesions are material entities located in the retina (vitreous haemorrhage in the vitreous body); venous beading and retinal thickening are qualities of a vein and of the macula~\cite{arp2015}. A finding value such as ``2 to 7 microaneurysms'' is a measurement datum about the eye, so each data property abbreviates an information entity about an eye~\cite{ceusters2012}, and a grade assignment is a data item about an eye made under a reading. A scheme is a plan specification that is part of its published document, and a reading is a directive information entity that precisifies a scheme. OWL has no standpoint modalities, so each grade becomes one class per reading, and ``under every reading'' is the intersection of those classes.

The vocabulary table has \AlnRows{} rows. \AlnSkos{} are SKOS mappings from \AlnItems{} of our items to \AlnTargetIds{} distinct terms: \AlnUBERON{} to UBERON~\cite{mungall2012}, \AlnHP{} to the Human Phenotype Ontology~\cite{gargano2024}, \AlnMONDO{} to Mondo~\cite{vasilevsky2026}, \AlnDOID{} to the Disease Ontology~\cite{schriml2022} and \AlnSNOMED{} to SNOMED CT~\cite{snomedct}; across all targets, \AlnExact{} are exact matches, \AlnClose{} close, \AlnBroad{} broad and \AlnNarrow{} narrow. For SNOMED CT we use only the \AlnSNOMEDConcepts{} concept identifiers and the names the label check needs (release \MapNetSnomedRelease). The Human Phenotype Ontology lacks venous beading, IRMA, venous loops and circinate exudates, which map to a parent. The \AlnNone{} remaining rows record that no suitable term exists, and they are informative: the pooled counts behind the 4-2-1 rule (``more than 20'' microaneurysms and haemorrhages per quadrant), the pooled distance of the nearest red lesion, a known non-oedema cause of reduced vision, the grader's uncertainty and every absence class (``no DME'', M0, ``DME absent'') have no standard term. The texts build their categories from exactly these thresholds and absences, where the gaps and overlaps of Section~\ref{sec:defects} arise.

Every image-visible variable (\AlnImageVars{}) and every grade or row class (\AlnBaseClasses{}) has a row. HermiT finds the module consistent with no unsatisfiable class on the base signature, and consistent, with the eye and feature classes satisfiable, on the union with the complete base ontology; a negative control that places a grade class under IAO data entity comes out unsatisfiable, as BFO's disjointness of independent and generically dependent continuants requires. Algorithm~\ref{alg:prune} gives the signature pruning and why it is sound. An online check re-fetches all \MapNetRows{} identifier rows, the \AlnSkos{} mappings and the \MapNetAlignTerms{} BFO and IAO terms the alignment module uses (\MapNetIds{} distinct identifiers), from OLS4, the BFO release and a SNOMED CT terminology server, and fails on a changed label or an obsolete or inactive term; on the last run every label was unchanged.

\section{How the claims are checked}
\label{sec:artefact}

Every category, table row and reading value (\NProvItems{} elements) is linked to at least one of \NProvQuotes{} verbatim quotes from \NProvSources{} source files; a checker confirms that each quote occurs in its source up to whitespace and hyphenation, that every element derived from the rules is covered, and that every action a quote states is the action the rules assign (Algorithm~\ref{alg:prov}). The check confirms that each quote exists and that the actions agree; whether a rule reads its quote correctly remains a matter of comparing the two, which the links make possible for each element. Building these links exposed errors in an earlier version of the transcription, among them three rows of the Italian table; they were corrected before any reported result was computed, and the language-model calls made on that version were archived unscored (Section~\ref{sec:llm}). Three engines share the transcription but not the reasoning: the enumerator, an SMT encoding in Z3, and the OWL~2 ontology, in which a gap is a satisfiable complement and an empty grade an unsatisfiable class. The enumerator and Z3 disagree on \NMismatch{} claims. HermiT is posed \NOwlQueries{} satisfiability questions, \NOwlControls{} of them negative controls that must be unsatisfiable; \OwlResult. The generation step of the build asserts \NClaims{} qualitative claims and stops the build if any fails. A sabotage suite of \NSabotage{} cases, from altering one character of a quote to misplacing an anatomical class in BFO, tests that each gate can fail: each sabotaged copy must fail with the expected message while an untouched copy passes (Algorithm~\ref{alg:sabotage}). All \NRefsChecked{} bibliography entries are checked against Crossref or their URL. Appendix~\ref{app:algs} gives each procedure as pseudocode.

\begin{table}[t]
\caption{Item-level results per model and condition (L: LITERAL, S: SCREENER). Repair: a single category on an item that licenses none or several; none and several: the defect reported as such. Each cell gives the count of items; the last column is the exact McNemar $p$ for repair, LITERAL against SCREENER.}
\label{tab:grid}
\footnotesize
\centering
\setlength{\tabcolsep}{4pt}
\begin{tabular}{@{}llccccccccccc@{}}
\toprule
& & \multicolumn{2}{c}{control correct} & \multicolumn{2}{c}{gap repair} & \multicolumn{2}{c}{overlap repair} & \multicolumn{2}{c}{gap: none} & \multicolumn{2}{c}{overlap: several} & \\
\cmidrule(lr){3-4}\cmidrule(lr){5-6}\cmidrule(lr){7-8}\cmidrule(lr){9-10}\cmidrule(lr){11-12}
model & thinking & L & S & L & S & L & S & L & S & L & S & $p$ \\
& & \multicolumn{2}{c}{of \GridCtlN} & \multicolumn{2}{c}{of \GridGapN} & \multicolumn{2}{c}{of \GridOvlN} & \multicolumn{2}{c}{of \GridGapN} & \multicolumn{2}{c}{of \GridOvlN} & \\
\midrule
\LlmModel{} & off & \QwenOffCtlLit & \QwenOffCtlScr & \QwenOffGapLit & \QwenOffGapScr & \QwenOffOvlLit & \QwenOffOvlScr & \QwenOffNoneLit & \QwenOffNoneScr & \QwenOffSevLit & \QwenOffSevScr & \QwenOffMcP \\
\LlmModel{} & on & \QwenOnCtlLit & \QwenOnCtlScr & \QwenOnGapLit & \QwenOnGapScr & \QwenOnOvlLit & \QwenOnOvlScr & \QwenOnNoneLit & \QwenOnNoneScr & \QwenOnSevLit & \QwenOnSevScr & \QwenOnMcP \\
\HaikuName & off & \HaikuCtlLit & \HaikuCtlScr & \HaikuGapLit & \HaikuGapScr & \HaikuOvlLit & \HaikuOvlScr & \HaikuNoneLit & \HaikuNoneScr & \HaikuSevLit & \HaikuSevScr & \HaikuMcP \\
\HaikuName & on & \HaikuThCtlLit & \HaikuThCtlScr & \HaikuThGapLit & \HaikuThGapScr & \HaikuThOvlLit & \HaikuThOvlScr & \HaikuThNoneLit & \HaikuThNoneScr & \HaikuThSevLit & \HaikuThSevScr & \HaikuThMcP \\
\SonnetName & on & \SonnetCtlLit & \SonnetCtlScr & \SonnetGapLit & \SonnetGapScr & \SonnetOvlLit & \SonnetOvlScr & \SonnetNoneLit & \SonnetNoneScr & \SonnetSevLit & \SonnetSevScr & \SonnetMcP \\
\OpusName & on & \OpusCtlLit & \OpusCtlScr & \OpusGapLit & \OpusGapScr & \OpusOvlLit & \OpusOvlScr & \OpusNoneLit & \OpusNoneScr & \OpusSevLit & \OpusSevScr & \OpusMcP \\
\bottomrule
\end{tabular}
\end{table}

\section{Do language models repair defective categories?}
\label{sec:llm}

\paragraph{Design.} We asked whether a language model applying a defective text reports the defect or repairs it silently. The study was pre-registered before any model call; \PreregStatement. Each item (a text, an axis and an eye in a fixed English template) instantiates one named defect or control class; the \NLlmItems{} items cover \NLlmDefects{} defects and \NLlmControls{} control classes in the \NLlmTexts{} texts (Algorithm~\ref{alg:llmitems}). Control items have one licensed category under every reading, gap items none, overlap items two or more, and reading-dependent items a set that varies with the reading, so a single category on a gap or overlap item is a silent repair. The model sees the verbatim text (Italian stays Italian) and may answer ``none'' or ``several''. The conditions differ only in their first sentence: LITERAL asks for the text to be applied exactly as written, SCREENER casts the model as an experienced grader. The primary unit is the defect, which has a property when a strict majority of its items have it; the primary test is an exact McNemar test of repair between conditions.

\paragraph{Models and amendments.} The registered run used \LlmModel{}, served locally with reasoning off. A first amendment, \AmendOneStatement, added \ClaudeModelsList{} through the Claude Code command-line interface (version \ClaudeCLIVersion) with the same items, prompts, scorer and decision rules; the interface exposes no temperature, so each Claude model answers once per item and condition at its default sampling. A check of the raw replies showed that this run kept only the last text block of some replies and lost others to thinking under the output cap, so it was archived. A second amendment, \AmendTwoStatement, re-ran all \ClaudeVTwoCalls{} Claude calls collecting every text block under a larger cap, and registered a hypothesis (H7) that deliberation reduces repair, with a reasoning-on arm for \LlmModel{}. A third amendment, \AmendThreeStatement, added \HaikuName{} with thinking on, to test H7 within a Claude model as well. A fourth amendment, \AmendFourStatement, completed the higher-temperature samples and the reasoning-on arm of \LlmModel{} on the same server, with nothing else changed. Two errata were corrected in code, without new calls, and the stored replies re-parsed: the canary's parser read the string ``false'' as a marker, and a paired test with no pairs returned ``failed'' instead of ``not testable''.

\paragraph{Isolation.} The calling machine holds the authors' own assistant configuration, so each call runs without tools, settings, plugins, hooks or memory in a fresh empty directory, and a canary established isolation first. It asked each of the \CanaryModels{} models for every instruction received and a detector whether any of \CanaryMarkers{} marker phrases from the authors' global instructions were present, with a positive control that planted them in the system prompt. Isolation required that no marker appear, the detector deny every marker, the control fire, and the minimal call's input stay under half of what planting adds. With the corrected parser, the canaries of both amendments prove isolation for all \CanaryModels{} models: minimal calls used \CanaryMinInLo{} to \CanaryMinInHi{} input tokens, while planting added \CanaryAddLo{} to \CanaryAddHi{}.

\begin{figure}[t]
\centering
\begin{minipage}[c]{0.3\linewidth}\centering
\includegraphics[width=0.85\linewidth]{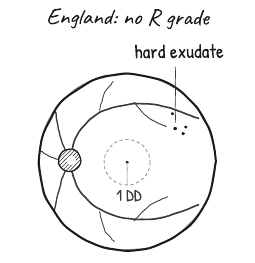}
\end{minipage}\hfill
\begin{minipage}[c]{0.66\linewidth}\centering\small
\input{generated/fig_repair.tex}
\end{minipage}
\caption{What each model answered on one gap item, by condition. \protect\input{generated/fig_repair_caption.tex}}
\label{fig:repair}
\end{figure}

\paragraph{Registered run.} \LlmModel{} behaved like the charitable reader the texts rely on: it answered every control item correctly, yet returned a single category on most gap and overlap items under either instruction (Table~\ref{tab:grid}, Figure~\ref{fig:repair}). H1, that under SCREENER most gap items get a single category, \LlmVerdictHOneItem{} at item level and was \LlmVerdictHOneDef{} at defect level, where England's gap counts as unrepaired under SCREENER only because \LlmENScrGapValid{} of its \LlmENScrGapN{} replies could not be parsed. The instruction made no difference to repair ($p=\LlmMcDefP$ over \LlmMcDefN{} defects). H2, that under LITERAL reports of ``none'' on gaps rise but stay below accuracy on controls, was \LlmVerdictHTwo{}: they stayed far below (Fisher $p=\LlmFisherP$), but the rise was \LlmVerdictHTwoA{}. H3, that under 20\% of overlap items are reported as ``several'', was \LlmVerdictHThree{}; H4, that answers on reading-dependent items concentrate on one reading, \LlmVerdictHFour{}; and the exploratory H5, of no difference in repair between the 2017 and 2003 texts on shared defects, fitted the data (every $p\ge\LlmHFiveMinP$). Over the temperature-0 answer and three samples at temperature 0.7, \LlmModel{} gave the same valid answer on \LlmConsLitK{} of \LlmConsN{} items under LITERAL and \LlmConsScrK{} under SCREENER.

\paragraph{Across models.} \HaikuName{} with thinking off behaved like \LlmModel{}: it repaired most gap and overlap items in both conditions, and the instruction to read literally made no significant difference. \SonnetName{} and \OpusName{}, which ran with thinking (\SonnetThinkTok{} and \OpusThinkTok{} thinking tokens in all), reported defects far more often: \OpusName{} repaired no gap item in either condition and answered ``none'' to every one, and both repaired significantly less under LITERAL than under SCREENER. Of H1 to H4, only the failure of H4 replicates in every Claude model (the exploratory H6), with the \ClaudeVTwoParseFail{} unparseable Claude replies, all from \SonnetName{}, counted as failures.

\paragraph{Deliberation within a model.} The two models that report defects both ran with thinking, which the interface keeps on for them; the provider's API has no setting that turns it off for \OpusName{}, and for \SonnetName{} only one outside the interface~\cite{anthropic_thinking2026}. H7 is therefore tested within two models whose reasoning can be switched both ways. In \LlmModel{}, reasoning lowered repair on the \WQLitPairs{} paired repair-eligible items under both instructions: it fell in \WQLitOff{} pairs and rose in \WQLitOn{} under LITERAL ($p=\WQLitP$), and fell in \WQScrOff{} and rose in \WQScrOn{} under SCREENER ($p=\WQScrP$), so H7 \WQVerdictItem{} at item level; at defect level it \WQDefVerdictLit{} under LITERAL (\WQDefLitOff{} against \WQDefLitOn{}, $p=\WQDefLitP$) and was \WQDefVerdictScr{} under SCREENER ($p=\WQDefScrP$). Counting the \QwenThinkTruncElig{} repair-eligible replies cut at the 8{,}000-token cap as repairs rather than failures still gives $p=\WQWorstLitP$ and $p=\WQWorstScrP$. In \HaikuName{}, all \HaikuThinkCalls{} calls used thinking (at least \HaikuThinkMin{} tokens, median \HaikuThinkMedian{}); thinking lowered gap and overlap repair and raised reports of ``none'' and ``several'' and control accuracy in both conditions (Table~\ref{tab:grid}), with repair falling in \WHLitOff{} pairs and rising in \WHLitOn{} under LITERAL ($p=\WHLitP$) and falling in \WHScrOff{} and rising in \WHScrOn{} under SCREENER ($p=\WHScrP$), so H7 as registered over both instructions was \WHVerdictItem{} at item level and \WHVerdictDefect{} at defect level; taken one instruction at a time, it \WHVerdictScr{} for SCREENER items and was \WHVerdictLit{} for LITERAL items. With thinking, both models still repaired some items, and \HaikuName{} most gap items, far more than \SonnetName{} or \OpusName{}, so deliberation alone did not make them literal readers.

\section{Discussion and conclusion}

The defects fall into two kinds. The first a competent reader repairs without notice, as worst-grade precedence absorbs overlaps: these are where rule-based categories are used like prototype categories~\cite{rosch1975,ashby2005}, and they fail wherever the text is executed rather than read. The second kind survives every reading: England and Italy order some eyes in opposite ways on what the camera sees, the English maculopathy grade turns on an unstated precedence, and the 2017 tabulation of the international scale is not the 2003 scale. These belong to the texts, not their readers. ``ICDR grade'' names a family of category systems, and pooling labels across them is an alignment problem in which non-functional crosswalks play the part of incoherent mappings~\cite{meilicke2009,jimenezruiz2011}. Standard terminologies name the lesions and disorders but not the thresholds and absences from which the texts build their categories, so a mapping to them records what a grade is about without recording what it says. That labels inherit their text was a conjecture; Section~\ref{sec:real} tests it for referral, where classifiers trained on English, ICO and Italian labels learn measurably different decisions and side with their own text where the labels differ.

The real-eye results sharpen the point of Section~\ref{sec:defects}. On annotated photographs, England's referral decisions part from the ICO and Italian ones only where the macula is assumed not thickened, and leaving thickening free removes every such disagreement, in the frequencies and in the training labels alike. A photograph-based grader trained on English labels therefore inherits a decision about something the photograph does not record. Whoever pools labels across these texts, or trains on one and deploys under another, needs to know which assumption about the unseen findings the labels carry.

The engines share one transcription, so an error in it would carry into every result; its verbatim quotes make each element checkable against the source wording, the reference the claims are about. A defect under every reading holds for the readings we admit, by the criterion of Section~\ref{sec:formal}, and each reading parameter is stated so that it can be contested. We formalise published text, not software. Every inversion we report is between actions whose order is not in doubt, so none depends on the common action scale. The frequencies of Section~\ref{sec:real} are for hospital images enriched for disease. The language-model study rests on \GridGapN{} gap and \GridOvlN{} overlap items, with one sample per Claude model and condition at default sampling; its exact tests compare conditions and arms on those items. The models that answered without deliberating repaired silently even when told to read literally, while \SonnetName{} and \OpusName{} reported defects far more often. Deliberation reduced repair within \LlmModel{} and \HaikuName{} but did not remove it, so deliberation alone made neither a literal reader.

\section*{Data}
All photographs come from public datasets released for research and were used under their licences: IDRiD under Creative Commons Attribution 4.0, the DDR lesion-segmentation subset under Creative Commons Attribution-NonCommercial-ShareAlike 4.0, and DIARETDB1 under its academic licence. The images are de-identified, we redistribute none of them, no new data were collected and no person took part in the study. The grading texts are public documents, quoted for analysis.

\section*{Declaration on Generative AI}
The authors used Claude (Anthropic) to draft and edit text, to write and test the formalisation, detection, frequency and training code, and to check references against Crossref. \LlmModel{} and \ClaudeModelsList{} were subjects of the study in Section~\ref{sec:llm}, called in isolation from the authoring sessions. The authors reviewed and edited all content and take full responsibility for the publication.

\bibliography{refs}

\appendix
\section{Algorithms}
\label{app:algs}

{\setlength{\intextsep}{4pt plus 2pt minus 2pt}\setlength{\abovecaptionskip}{2pt}
\input{appendix_algorithms.tex}}

\end{document}

%% file: generated/numbers.tex
\newcommand{\SpaceSize}{1{,}425{,}807{,}360}
\newcommand{\ImageSpaceSize}{79{,}211{,}520}
\newcommand{\NVars}{24}

\newcommand{\NPEN}{12}
\newcommand{\NPIC}{24}
\newcommand{\NPIT}{48}
\newcommand{\NPIThree}{16}

\newcommand{\ICOvlDRReadings}{16}

\newcommand{\NMismatch}{0}
\newcommand{\NConstraints}{16}
\newcommand{\NOwlQueries}{29}
\newcommand{\NOwlControls}{four}

\newcommand{\IdMatched}{60}

\newcommand{\DmeContra}{two}
\newcommand{\ENICdis}{two}

\newcommand{\NDraws}{1{,}000}

\newcommand{\FigXDesc}{eight or more microaneurysms, more than 20 microaneurysms/haemorrhages in 1 quadrant}
\newcommand{\FigXEN}{recall}
\newcommand{\FigXIC}{recall}
\newcommand{\FigXIT}{recall}
\newcommand{\FigYDesc}{a venous loop}
\newcommand{\FigYEN}{rescreen}
\newcommand{\FigYIC}{no grade}
\newcommand{\FigYIT}{refer}
\newcommand{\NProvItems}{75}
\newcommand{\NProvQuotes}{90}

\newcommand{\NProvSources}{five}
\newcommand{\NSabotage}{ten}
\newcommand{\NRefsChecked}{41}
\newcommand{\PreregStatement}{it was registered at 2026-09-30 22:04:35 UTC (SHA-256 prefix \texttt{eb0c01077349}); an earlier version was withdrawn, its calls archived unscored, when the transcription was corrected}
\newcommand{\NLlmItems}{140}
\newcommand{\NLlmDefects}{31}
\newcommand{\NLlmControls}{28}

\newcommand{\NLlmTexts}{four}
\newcommand{\LlmModel}{Qwen3.8-Flash-Next}

\newcommand{\LlmMcDefP}{1}
\newcommand{\LlmMcDefN}{13}

\newcommand{\LlmFisherP}{1.5\times 10^{-11}}

\newcommand{\LlmHFiveMinP}{0.125}
\newcommand{\LlmConsLitK}{105}
\newcommand{\LlmConsScrK}{85}
\newcommand{\LlmConsN}{140}
\newcommand{\LlmENScrGapValid}{two}
\newcommand{\LlmENScrGapN}{four}
\newcommand{\LlmVerdictHOneItem}{held}
\newcommand{\LlmVerdictHOneDef}{inconclusive}
\newcommand{\LlmVerdictHTwo}{inconclusive}
\newcommand{\LlmVerdictHTwoA}{inconclusive}

\newcommand{\LlmVerdictHThree}{inconclusive}
\newcommand{\LlmVerdictHFour}{failed}
\newcommand{\OwlResult}{it agreed with the SMT encoding on all 29 of them}
\newcommand{\AlnRows}{165}
\newcommand{\AlnSkos}{153}
\newcommand{\AlnNone}{twelve}
\newcommand{\AlnItems}{79}
\newcommand{\AlnTargetIds}{60}
\newcommand{\AlnUBERON}{eight}
\newcommand{\AlnHP}{20}
\newcommand{\AlnMONDO}{25}
\newcommand{\AlnDOID}{25}
\newcommand{\AlnSNOMED}{75}
\newcommand{\AlnSNOMEDConcepts}{28}
\newcommand{\AlnExact}{18}
\newcommand{\AlnClose}{72}
\newcommand{\AlnBroad}{61}
\newcommand{\AlnNarrow}{two}
\newcommand{\AlnImageVars}{20}
\newcommand{\AlnBaseClasses}{60}

\newcommand{\MapNetRows}{164}
\newcommand{\MapNetIds}{71}
\newcommand{\MapNetSnomedRelease}{2026-07-01}
\newcommand{\CanaryModels}{three}
\newcommand{\CanaryMarkers}{five}
\newcommand{\CanaryMinInLo}{388}
\newcommand{\CanaryMinInHi}{451}
\newcommand{\CanaryAddLo}{9{,}966}
\newcommand{\CanaryAddHi}{36{,}565}
\newcommand{\ClaudeCLIVersion}{2.1.284}
\newcommand{\ClaudeModelsList}{Claude Haiku 4.5, Claude Sonnet 5.5 and Claude Opus 5.5}
\newcommand{\HaikuName}{Claude Haiku 4.5}
\newcommand{\SonnetName}{Claude Sonnet 5.5}
\newcommand{\OpusName}{Claude Opus 5.5}
\newcommand{\AmendOneStatement}{registered at 2026-10-01 13:00:04 UTC (SHA-256 prefix \texttt{b8e9d4a91a31})}
\newcommand{\AmendTwoStatement}{registered at 2026-10-01 13:33:45 UTC (SHA-256 prefix \texttt{2391fd4e62b8})}
\newcommand{\QwenOffCtlLit}{45}
\newcommand{\QwenOffGapLit}{12}
\newcommand{\QwenOffOvlLit}{22}
\newcommand{\QwenOffNoneLit}{1}
\newcommand{\QwenOffSevLit}{5}
\newcommand{\QwenOffCtlScr}{45}
\newcommand{\QwenOffGapScr}{11}
\newcommand{\QwenOffOvlScr}{23}
\newcommand{\QwenOffNoneScr}{0}
\newcommand{\QwenOffSevScr}{4}
\newcommand{\QwenOffMcP}{1}
\newcommand{\QwenOnCtlLit}{44}
\newcommand{\QwenOnGapLit}{5}
\newcommand{\QwenOnOvlLit}{7}
\newcommand{\QwenOnNoneLit}{7}
\newcommand{\QwenOnSevLit}{18}
\newcommand{\QwenOnCtlScr}{44}
\newcommand{\QwenOnGapScr}{8}
\newcommand{\QwenOnOvlScr}{12}
\newcommand{\QwenOnNoneScr}{5}
\newcommand{\QwenOnSevScr}{10}
\newcommand{\QwenOnMcP}{0.0078}
\newcommand{\HaikuCtlLit}{37}
\newcommand{\HaikuGapLit}{12}
\newcommand{\HaikuOvlLit}{25}
\newcommand{\HaikuNoneLit}{1}
\newcommand{\HaikuSevLit}{2}
\newcommand{\HaikuCtlScr}{40}
\newcommand{\HaikuGapScr}{13}
\newcommand{\HaikuOvlScr}{27}
\newcommand{\HaikuNoneScr}{0}
\newcommand{\HaikuSevScr}{0}
\newcommand{\HaikuMcP}{0.25}
\newcommand{\SonnetCtlLit}{44}
\newcommand{\SonnetGapLit}{2}
\newcommand{\SonnetOvlLit}{9}
\newcommand{\SonnetNoneLit}{11}
\newcommand{\SonnetSevLit}{16}
\newcommand{\SonnetCtlScr}{45}
\newcommand{\SonnetGapScr}{5}
\newcommand{\SonnetOvlScr}{18}
\newcommand{\SonnetNoneScr}{8}
\newcommand{\SonnetSevScr}{8}
\newcommand{\SonnetMcP}{0.00049}
\newcommand{\OpusCtlLit}{40}
\newcommand{\OpusGapLit}{0}
\newcommand{\OpusOvlLit}{6}
\newcommand{\OpusNoneLit}{13}
\newcommand{\OpusSevLit}{21}
\newcommand{\OpusCtlScr}{41}
\newcommand{\OpusGapScr}{0}
\newcommand{\OpusOvlScr}{13}
\newcommand{\OpusNoneScr}{13}
\newcommand{\OpusSevScr}{13}
\newcommand{\OpusMcP}{0.039}
\newcommand{\GridCtlN}{45}
\newcommand{\GridGapN}{13}
\newcommand{\GridOvlN}{27}
\newcommand{\ClaudeVTwoCalls}{840}
\newcommand{\ClaudeVTwoParseFail}{two}
\newcommand{\SonnetThinkTok}{60{,}125}
\newcommand{\OpusThinkTok}{70{,}371}

\newcommand{\WQLitOff}{23}
\newcommand{\WQLitOn}{1}
\newcommand{\WQLitP}{3.0\times 10^{-6}}
\newcommand{\WQLitPairs}{40}
\newcommand{\WQDefLitOff}{8}
\newcommand{\WQDefLitOn}{0}
\newcommand{\WQDefLitP}{0.0078}

\newcommand{\WQDefVerdictLit}{held}
\newcommand{\WQScrOff}{17}
\newcommand{\WQScrOn}{3}
\newcommand{\WQScrP}{0.0026}

\newcommand{\WQDefScrP}{0.12}

\newcommand{\WQDefVerdictScr}{inconclusive}
\newcommand{\WQVerdictItem}{held}

\newcommand{\WQWorstLitP}{6.6\times 10^{-5}}
\newcommand{\WQWorstScrP}{0.035}

\newcommand{\QwenThinkTruncElig}{eight}
\newcommand{\AmendThreeStatement}{registered at 2026-10-01 20:25:31 UTC (SHA-256 prefix \texttt{95a0cc818d36})}
\newcommand{\AmendFourStatement}{registered at 2026-10-05 23:28:29 UTC (SHA-256 prefix \texttt{c7251b2dca07})}
\newcommand{\HaikuThinkMin}{277}
\newcommand{\HaikuThinkMedian}{1{,}172.5}
\newcommand{\HaikuThinkCalls}{280}
\newcommand{\HaikuThCtlLit}{44}
\newcommand{\HaikuThGapLit}{10}
\newcommand{\HaikuThOvlLit}{21}
\newcommand{\HaikuThNoneLit}{3}
\newcommand{\HaikuThSevLit}{6}
\newcommand{\WHLitOff}{8}
\newcommand{\WHLitOn}{2}
\newcommand{\WHLitP}{0.11}

\newcommand{\WHVerdictLit}{inconclusive}
\newcommand{\HaikuThCtlScr}{44}
\newcommand{\HaikuThGapScr}{11}
\newcommand{\HaikuThOvlScr}{18}
\newcommand{\HaikuThNoneScr}{2}
\newcommand{\HaikuThSevScr}{9}
\newcommand{\WHScrOff}{11}
\newcommand{\WHScrOn}{0}
\newcommand{\WHScrP}{0.00098}

\newcommand{\WHVerdictScr}{held}
\newcommand{\HaikuThMcP}{0.75}
\newcommand{\WHVerdictItem}{inconclusive}
\newcommand{\WHVerdictDefect}{inconclusive}
\newcommand{\DetGateSha}{\texttt{b9b1e9dd9b52}}

\newcommand{\DetArch}{ResNet-34}

\newcommand{\DetNTest}{103}
\newcommand{\DetNMasks}{81}
\newcommand{\DetFovMed}{0.045}
\newcommand{\DetFovP}{0.214}
\newcommand{\DetFovMedThr}{0.15}
\newcommand{\DetFovPThr}{$1/3$}
\newcommand{\DetDiscMed}{0.021}
\newcommand{\DetDiscThr}{0.15}
\newcommand{\DetDDMed}{1.8}
\newcommand{\DetDDThr}{10}
\newcommand{\DetEFov}{0.181}
\newcommand{\DetEDisc}{0.063}
\newcommand{\DetEDD}{5.3}

\newcommand{\DetCovN}{60}
\newcommand{\DetCovChecks}{600}

\newcommand{\DetCovNoMarginMiss}{twelve}

\newcommand{\DetDdrDDMed}{15}
\newcommand{\DetDdrDDP}{53}
\newcommand{\DetDdbDDMed}{15}
\newcommand{\DetDdbDDP}{38}
\newcommand{\FqIdN}{81}
\newcommand{\FqDdrN}{757}
\newcommand{\FqDdbN}{89}
\newcommand{\FqIdSixty}{60}

\newcommand{\FqICmacDdrLo}{251}
\newcommand{\FqICmacDdrHi}{478}

\newcommand{\FqICnoMA}{187}
\newcommand{\FqENgapDdr}{five}
\newcommand{\FqENgapDdb}{two}
\newcommand{\FqENovDdrLo}{149}
\newcommand{\FqENovDdrHi}{366}
\newcommand{\FqICovDdr}{88}
\newcommand{\FqICovDdb}{nine}
\newcommand{\FqITovDdrLo}{129}
\newcommand{\FqITovDdrHi}{321}
\newcommand{\FqDdrENICdisN}{153}
\newcommand{\FqDdrENICbothN}{570}
\newcommand{\FqDdrENICdisPct}{26.8}
\newcommand{\FqDdrENICdisLo}{23.4}
\newcommand{\FqDdrENICdisHi}{30.6}
\newcommand{\FqDdrENITdisN}{63}
\newcommand{\FqDdrENITbothN}{403}
\newcommand{\FqDdrICITdisN}{seven}
\newcommand{\FqDdrICITbothN}{279}

\newcommand{\FqPrevMA}{75.3}
\newcommand{\FqPrevHE}{79.4}
\newcommand{\FqPrevEX}{64.2}

\newcommand{\FqENgapDdrPct}{0.7}
\newcommand{\FqENgapDdrLo}{0.3}
\newcommand{\FqENgapDdrHi}{1.5}
\newcommand{\TrPreregUTC}{2026-10-01 22:48}
\newcommand{\TrPreregSha}{\texttt{f300e6038d3b}}
\newcommand{\TrArch}{ResNet-18}
\newcommand{\TrN}{310}
\newcommand{\TrNAll}{927}
\newcommand{\TrNId}{62}
\newcommand{\TrNDdr}{192}
\newcommand{\TrNDdb}{56}
\newcommand{\TrFolds}{five}
\newcommand{\TrPrevEN}{54.2}
\newcommand{\TrPrevIC}{71.9}
\newcommand{\TrPrevIT}{69.7}

\newcommand{\TrTripNone}{87}
\newcommand{\TrTripIC}{seven}
\newcommand{\TrTripICIT}{48}
\newcommand{\TrTripAll}{168}

\newcommand{\TrTestable}{four}

\newcommand{\TrHolmMax}{0.002}
\newcommand{\TrHolmMin}{4.8\times 10^{-7}}
\newcommand{\TrUnderN}{seven}
\newcommand{\TrUnderThr}{10}
\newcommand{\TrSideLo}{75}
\newcommand{\TrSideHi}{85}

\newcommand{\TrEightVerdict}{partially supported}
\newcommand{\TrTenVerdict}{partially supported}
\newcommand{\TrDevImgs}{seven}

\newcommand{\TfICgapId}{0.0}
\newcommand{\TfICgapDdr}{24.7}
\newcommand{\TfICgapDdb}{5.6}
\newcommand{\TfICgapDdrN}{187}
\newcommand{\TfDdrENICbothN}{147}
\newcommand{\TfDdrENITbothN}{136}
\newcommand{\TfDdrICITbothN}{279}
\newcommand{\TfDdrICITdisN}{seven}
\newcommand{\TrTfENICdis}{55}
\newcommand{\TrTfENITdis}{48}

\newcommand{\MapNetAlignTerms}{eleven}
\newcommand{\NAlgorithms}{seven}
\newcommand{\NClaims}{163}
\newcommand{\NMakeGates}{15}

%% file: generated/fig_inversion.tex
\begin{tikzpicture}[x=3.1cm,y=0.9cm,font=\small]
\node[anchor=east,text=gray] at (-0.15,-1) {no grade};
\draw[gray!30] (0,-1) -- (2,-1);
\node[anchor=east,text=gray] at (-0.15,0) {rescreen};
\draw[gray!30] (0,0) -- (2,0);
\node[anchor=east,text=gray] at (-0.15,1) {recall};
\draw[gray!30] (0,1) -- (2,1);
\node[anchor=east,text=gray] at (-0.15,2) {refer};
\draw[gray!30] (0,2) -- (2,2);
\node[anchor=east,text=gray] at (-0.15,3) {urgent};
\draw[gray!30] (0,3) -- (2,3);
\draw[gray!60] (0,-1.3) -- (0,3.3);
\node[anchor=north] at (0,-1.35) {England};
\draw[gray!60] (1,-1.3) -- (1,3.3);
\node[anchor=north] at (1,-1.35) {Italy};
\draw[gray!60] (2,-1.3) -- (2,3.3);
\node[anchor=north] at (2,-1.35) {ICDR};
\draw[black,thick] (0,1) -- (1,1) -- (2,1);
\fill (0,1) circle (2.2pt);
\fill (1,1) circle (2.2pt);
\fill (2,1) circle (2.2pt);
\node[anchor=west] at (2.08,1) {$x$};
\draw[black,thick,dashed] (0,0) -- (1,2) -- (2,-1);
\fill (0,0) circle (2.2pt);
\fill (1,2) circle (2.2pt);
\fill (2,-1) circle (2.2pt);
\node[anchor=west] at (2.08,-1) {$y$};
\end{tikzpicture}

%% file: generated/fig_inversion_caption.tex
$x$: 8+ microaneurysms, more than 20 microaneurysms/haemorrhages in 1 quadrant. $y$: a venous loop. No other findings.

%% file: generated/fig_gaps_caption.tex
England: hard exudate outside the macula, no R grade, under every reading. ICO 2017: cotton wool spot, no DR grade, under every reading. Italy: vitreous haemorrhage, no table row applies, under every reading. ICDR 2003: haemorrhage, no DR grade, only when the no-PDR clause is in the lead-in and moderate requires microaneurysms (4 of 16 readings; no eye has a gap under every reading).

%% file: generated/fig_versions.tex
\begin{tabular}{@{}l l >{\raggedright\arraybackslash}p{0.355\linewidth} >{\raggedright\arraybackslash}p{0.355\linewidth}@{}}
\toprule
eye & axis & ICDR 2003 & ICO 2017 \\
\midrule
\textit{a} & DR & \textbf{PDR}\newline{\footnotesize moderate requires microaneurysms (8 of 16)} & \textbf{no grade}\newline{\footnotesize PDR needs the severe features or PDR is read literally (16 of 24)} \\[2pt]
 &  & \textbf{moderate, PDR}\newline{\footnotesize moderate is anything beyond microaneurysms only (8 of 16)} & \textbf{PDR}\newline{\footnotesize PDR is any new vessel or vitreous haemorrhage (8 of 24)} \\
\midrule
\textit{b} & DME & \textbf{present}\newline{\footnotesize all 16 readings} & \textbf{no grade}\newline{\footnotesize all 24 readings} \\
\bottomrule
\end{tabular}

%% file: generated/fig_versions_caption.tex
Eye \textit{a}: new vessels. Eye \textit{b}: hard exudate 1/3 to 1 DD from the fovea. No other findings, no retinal thickening.

%% file: generated/fig_overlap.tex
\begin{tabular}{@{}r l l@{}}
\toprule
row & Italian Tab.~3 (verbatim; our English gloss below) & action \\
\midrule
\textcolor{black!45}{1} & \textcolor{black!45}{\footnotesize does not apply} & \textcolor{black!45}{rescreen} \\
$\blacktriangleright$\,\textbf{2} & \textbf{RD non proliferante lieve} & \textbf{rescreen} \\
 & {\footnotesize\itshape ``microaneurismi isolati'', screening a 12 mesi} & \\
 & {\footnotesize (mild non-proliferative DR: ``isolated microaneurysms'', screening at 12 months)} & \\[2pt]
$\blacktriangleright$\,\textbf{3} & \textbf{RD non proliferante moderata} & \textbf{recall} \\
 & {\footnotesize\itshape ``emorragie isolate e/o microaneurismi'', screening a 6-12 mesi} & \\
 & {\footnotesize (moderate non-proliferative DR: ``isolated haemorrhages and/or microaneurysms'', screening at 6-12 months)} & \\[2pt]
\textcolor{black!45}{4} & \textcolor{black!45}{\footnotesize does not apply} & \textcolor{black!45}{refer} \\
\textcolor{black!45}{5} & \textcolor{black!45}{\footnotesize does not apply} & \textcolor{black!45}{refer} \\
\textcolor{black!45}{6} & \textcolor{black!45}{\footnotesize does not apply} & \textcolor{black!45}{urgent} \\
\textcolor{black!45}{7} & \textcolor{black!45}{\footnotesize does not apply} & \textcolor{black!45}{urgent} \\
\bottomrule
\end{tabular}

%% file: generated/fig_overlap_caption.tex
The eye: microaneurysm, no other findings, no visual disturbance. Under every one of the 48 readings rows 2 and 3 of the Italian table apply, with actions rescreen and recall.

%% file: generated/tab_realeyes.tex
England & gap & every & 0.0 & 0.7 & 2.2 \\
 & overlap & every & 65.4--79.0 & 20.7--48.3 & 4.5--29.2 \\
ICO 2017 & gap & every & 97.5--100.0 & 48.5--70.1 & 33.7--44.9 \\
 & overlap & every & 0.0 & 11.6 & 10.1 \\
ICDR 2003 & gap & some & 0.0 & 24.6--24.7 & 5.6 \\
 & grade set varies &  & 100.0 & 99.9--100.0 & 96.6--100.0 \\
Italy & gap & some & 0.0--8.6 & 2.0--15.6 & 1.1--14.6 \\
 & overlap & every & 18.5--74.1 & 21.9--48.2 & 7.9--37.1 \\
 & row set varies &  & 100.0 & 76.4--96.3 & 55.1--61.8 \\
\midrule
ICO 2017, thickening free & gap & every & 0.0--100.0 & 24.7--70.1 & 5.6--44.9 \\
\bottomrule

%% file: generated/tab_train.tex
England & 92.3 (88.7--94.7) & 79.7 (74.8--83.8) & 81.9 (77.3--85.8) \\
ICO 2017 & 81.3 (76.6--85.2) & 91.9 (88.4--94.5) & 92.9 (89.5--95.3) \\
Italy & 85.2 (80.8--88.7) & 91.3 (87.6--93.9) & 92.9 (89.5--95.3) \\
\bottomrule

%% file: generated/fig_repair.tex
\begin{tabular}{@{}lcc@{}}
\toprule
 & \multicolumn{1}{c}{literal} & \multicolumn{1}{c}{screener} \\
\cmidrule(lr){2-2} \cmidrule(lr){3-3}
text licenses & \textbf{none} & \textbf{none} \\
\midrule
Qwen3.8-Flash-Next, no reasoning & R1 & R1 \\
Qwen3.8-Flash-Next, reasoning & R0 & R0 \\
Claude Haiku 4.5 & R0 & R0 \\
Claude Haiku 4.5, thinking & R0 & R0 \\
Claude Sonnet 5.5, thinking & \textbf{none} & \textbf{none} \\
Claude Opus 5.5, thinking & \textbf{none} & \textbf{none} \\
\bottomrule
\end{tabular}

%% file: generated/fig_repair_caption.tex
England, axis R: hard exudate outside the macula, no other findings. The text assigns no R grade under any of its 12 readings, so the licensed answer is none.

%% file: appendix_algorithms.tex
\begin{algorithm}[H]
\caption{Critical-value abstraction and exhaustive enumeration over the bins and constraints of the transcription.}
\label{alg:enum}
\begin{algorithmic}[1]
\Require bins $D_v$ of each $v\in V$ (one between consecutive thresholds of any text), constraints $\Gamma$, readings $P_S$ of each $S\in\mathcal{T}$; $V=L\cup E\cup N$, the lesion and lesion-distance variables other than exudates $L$, the exudate variables $E$ (presence, distance from the fovea, grouping, circinate pattern) and the non-image variables $N$; slice width $C$ (profiles per parallel task)
\Ensure the first witness of every claim key that some profile satisfies
\State $X_L\gets$ profiles over $L$ satisfying $\Gamma$; $X_R\gets$ profiles over $E\cup N$ satisfying $\Gamma$ \Comment{no constraint spans both; each built per connected component of $\Gamma$}
\For{each image key $k$ of $X_R$ and slice $X_i$ of $X_L$ of width $C$, as parallel tasks}
  \For{each $r\in X_R$ with image part $k$}
    \State evaluate each $S$ under each $p\in P_S$ on $X_i\times\{r\}$: gaps, overlaps, grades, rows, worst action $a^p_S$
    \State note the first witness of each defect, crosswalk cell, co-occurring pair of 2003 and 2017 grades, supervaluation key and jointly determinate action pair (Algorithm~\ref{alg:det})
  \EndFor
  \State note image dependence: the first $x$ whose action changes with the non-image part alone
\EndFor
\State merge task results in task order, keeping the first witness of each key; a grade or row never noted is empty
\end{algorithmic}
\end{algorithm}

\begin{algorithm}[H]
\caption{Minimal witness extraction with Z3. Supervaluation, pair and inversion queries minimise; per-reading queries use a plain solver.}
\label{alg:witness}
\begin{algorithmic}[1]
\Require query $\varphi(x^1,\dots,x^k)$ over $k$ profiles; bins $D_v$; constraints $\Gamma$; baseline $b$
\Ensure \textit{unsat}, or a witness with the fewest values away from $b$
\State for each $j\leq k$: fresh integer constants $x^j_v$, $v\in V$, with $\bigwedge_v\bigvee_{d\in D_v}(x^j_v=d)\wedge\bigwedge_{\gamma\in\Gamma}\gamma(x^j)$
\State assert these and $\varphi$ in a Z3 optimiser (or plain solver); if minimising, minimise $\sum_{j}\sum_{v}[x^j_v\neq b_v]$
\State \Return \textit{unsat} unless the check returns \textit{sat}; otherwise the model, completing any free constant
\end{algorithmic}
\end{algorithm}

\begin{algorithm}[H]
\caption{Supervaluated determinacy, structural disagreement and order inversion; the image-only queries are re-derived by a second enumeration. Queries are decided by Algorithm~\ref{alg:witness}.}
\label{alg:det}
\begin{algorithmic}[1]
\Require action texts $\mathcal{T}_a$ with readings $P_S$; $a^p_S(x)$ the worst action that applies, $\bot$ if some axis has none; referral threshold $\rho$; enumerated cell sets $K^{\mathrm{enum}}_{ST}$
\State $\textsc{Det}(S,x)\equiv a_S(x)\neq\bot\wedge\bigwedge_{p\in P_S}a^p_S(x)=a_S(x)$, with $a_S=a^{p_0}_S$ for the first reading $p_0$
\State for each text: decide a gap under every reading, a gap under some but not every reading, and (action texts) no gap but $\neg\textsc{Det}$
\For{each pair $(S,T)$ of action texts}
  \State $K_{ST}\gets$ the action pairs $(u,v)$ with $\textsc{Det}(S,x)\wedge\textsc{Det}(T,x)\wedge a_S(x)=u\wedge a_T(x)=v$ satisfiable; disagreement if some $u\neq v$
  \State inversion: two profiles, all four actions determinate, $a_S(x)>a_S(y)\wedge a_T(x)<a_T(y)$
  \State record a mismatch if $K_{ST}\neq K^{\mathrm{enum}}_{ST}$ or the inversion differs from the one read off $K^{\mathrm{enum}}_{ST}$
\EndFor
\State image only: fix the non-image variables at baseline, decide inversion and referral disagreement ($a\geq\rho$), and compare with enumeration of the restricted space; fail on any mismatch
\end{algorithmic}
\end{algorithm}

\begin{algorithm}[H]
\caption{Provenance check over the verbatim quotes and the source defects.}
\label{alg:prov}
\begin{algorithmic}[1]
\Require verbatim sources; items with quotes and an optional stated action; window $W$; minimum fragment length $m$
\State \textsc{Norm}: Unicode NFKC, Unicode hyphens to ASCII, join hyphenated line breaks, collapse whitespace, drop hyphens between letters
\State \textsc{Check}$(q)$: restrict the normalised source to $q$'s section if it has one (its start must occur once); a single quote must occur verbatim; a fragmented quote needs $\geq 2$ fragments of length $\geq m$, in order, within $W$ characters
\State $\mathit{Need}\gets$ every grade, table row and reading value of each text, with the actions the transcription assigns it over all readings
\State fail if an item is missing from $\mathit{Need}$ or has no quote, if any \textsc{Check} fails, or if a stated action $k$ differs from $\mathit{Need}[e]=\{k\}$
\end{algorithmic}
\end{algorithm}

\begin{algorithm}[H]
\caption{Sabotage protocol. Each run uses a fresh temporary copy of the project, with data and the virtual environment symlinked, so the real files are never modified.}
\label{alg:sabotage}
\begin{algorithmic}[1]
\Require cases $(i,t,\sigma,g)$: identifier, make target of the gate, edit, expected signature; network gates $N$
\For{each case $(i,t,\sigma,g)$}
  \State run $t$ on an untouched copy (once per target); if $t\in N$ and the network is unreachable, report $i$ skipped; if it fails, $i$ fails
  \State apply $\sigma$ to a fresh copy (each anchor must occur once) and run $t$; $i$ fired only if $t$ fails and its output contains $g$
\EndFor
\State \Return failure if any case did not fire and was not skipped
\end{algorithmic}
\end{algorithm}

\begin{algorithm}[H]
\caption{Relevance pruning for the HermiT checks and the alignment gate. Sound because every constraint outside $\Gamma_R$ mentions only variables outside $R$ and the baseline $\beta$ satisfies every constraint, so $\mathrm{SAT}(Q\wedge\Gamma)\Leftrightarrow\mathrm{SAT}(Q\wedge\Gamma_R)$.}
\label{alg:prune}
\begin{algorithmic}[1]
\Require variables $V$, domains $D_v$, constraints $\Gamma$, baseline $\beta$, queries $\mathcal{Q}$ (defects and negative controls)
\State $R(Q)\gets$ the closure of $\mathrm{vars}(Q)$ under sharing a constraint; $\Gamma_R\gets$ the constraints over $R$
\State verify with Z3 that $\beta$ satisfies $\Gamma$ and that $\mathrm{SAT}(Q\wedge\Gamma)=\mathrm{SAT}(Q\wedge\Gamma_R)$ for every $Q$; abort otherwise
\State for each $Q$, one HermiT process: Eye with a functional data property per $v\in R$, $\Gamma_R$ as axioms, $\mathit{Query}\equiv\mathit{Eye}\sqcap Q$; require every negative control to be unsatisfiable
\State alignment: classify module $+$ BFO $+$ IAO $+$ base signature, check the complete union with \texttt{-k}, and reject a grade class placed under IAO data entity
\end{algorithmic}
\end{algorithm}

\begin{algorithm}[H]
\caption{Construction and scoring of the language-model items. $F_p$ is the set of labels on the item's axis whose rule holds under reading $p$. A reply that fails to parse counts as a failure on every metric.}
\label{alg:llmitems}
\begin{algorithmic}[1]
\Require defects, each with text $s$, axis, stratum (gap, overlap, control or reading-dependent) and variants $\varphi_k$ (alternative formulas that witness the same defect); readings $P_s$; baseline $\beta$
\For{each defect and variant $\varphi_k$}
  \State $w\gets$ the Z3 profile closest to $\beta$ satisfying the domains, $\Gamma$, $\varphi_k$ and the stratum goal, unused in $s$
  \State $F_p\gets$ labels holding at $w$ under $p$, by numpy and checked against Z3; assert the stratum: reading-dependent if $F_p$ varies, else gap, control or overlap as $|F_p|$ is $0$, $1$ or more
\EndFor
\State \textsc{Score}: $G\gets$ the returned labels; licensed iff $G=F_p$ for some $p$; repair iff $|G|=1$ and $|F_p|\neq 1$ for every $p$; a reported gap or overlap needs status none with $G=\emptyset$, or several with $|G|\geq 2$
\State defect level: a defect has a property iff a strict majority of its items have it
\end{algorithmic}
\end{algorithm}